\pdfoutput=1

\documentclass[11pt]{article}

\usepackage[preprint]{acl}

\usepackage{times}
\usepackage{latexsym}
\usepackage[T1]{fontenc}
\usepackage[utf8]{inputenc}
\usepackage{microtype}
\usepackage{inconsolata}
\usepackage{graphicx}
\usepackage{amsmath}

\usepackage{booktabs}
\usepackage{multirow}
\usepackage{lscape}
\usepackage{tcolorbox}
\usepackage{enumitem}
\usepackage{arydshln}
\usepackage{xcolor}
\usepackage[flushleft]{threeparttable}

\usepackage{geometry}
\usepackage{longtable}
\usepackage{adjustbox}
\usepackage{array}

\newcommand{\method}{\textsc{Typed-RAG}}
\title{\method: Type-Aware Decomposition of Non-Factoid Questions \\ for Retrieval-Augmented Generation}

\author{
  DongGeon Lee\textsuperscript{\rm 1}\Thanks{These authors contributed equally to this work.}
  \quad
  Ahjeong Park\textsuperscript{\rm 2}$^*$
  \quad
  Hyeri Lee\textsuperscript{\rm 3}
  \quad
  Hyeonseo Nam\textsuperscript{\rm 4}
  \quad
  Yunho Maeng\textsuperscript{\rm 5, \rm 6}
  \\
  $^1$POSTECH
  \quad
  $^2$Sookmyung Women's University
  \\
  $^3$Independent Researcher
  \quad
  $^4$KT 
  \quad
  $^5$Ewha Womans University
  \\
  $^6$LLM Experimental Lab, MODULABS
  \\
   \small{
   \textbf{Correspondence:} \href{mailto:yunhomaeng@ewha.ac.kr}{yunhomaeng@ewha.ac.kr}
 }
}

\begin{document}
\maketitle

\begin{abstract}

Addressing non-factoid question answering (NFQA) remains challenging due to its open-ended nature, diverse user intents, and need for multi-aspect reasoning. 
These characteristics often reveal the limitations of conventional retrieval-augmented generation (RAG) approaches. 
To overcome these challenges, we propose \method, a framework for type-aware decomposition of non-factoid questions (NFQs) within the RAG paradigm. 
Specifically, \method\ first classifies an NFQ into a predefined type (e.g., Debate, Experience, Comparison).
It then decomposes the question into focused sub-queries, each focusing on a single aspect. 
This decomposition enhances both retrieval relevance and answer quality.
By combining the results of these sub-queries, \method\ produces more informative and contextually aligned responses. 
Additionally, we construct Wiki-NFQA, a benchmark dataset for NFQA covering a wide range of NFQ types.
Experiments show that \method\ consistently outperforms existing QA approaches based on LLMs or RAG methods, validating the effectiveness of type-aware decomposition for improving both retrieval quality and answer generation in NFQA.  
Our code and dataset are available on \url{https://github.com/TeamNLP/Typed-RAG}.

\end{abstract}

\section{Introduction}
Traditional and current question answering (QA) systems \cite{rajpurkar-etal-2016-squad, Peters2018ELMO, Lewis2020RAG, Ouyang2022InstructGPT, Zhang2024Enhancing} have primarily addressed \textbf{factoid} questions---queries seeking specific, verifiable information that yield concise, objective answers like \textit{``When was Google founded?''}. 
However, real-world information needs extend far beyond such simple factual queries.

\begin{figure}[t!]
    \centering
    \includegraphics[width=\linewidth]{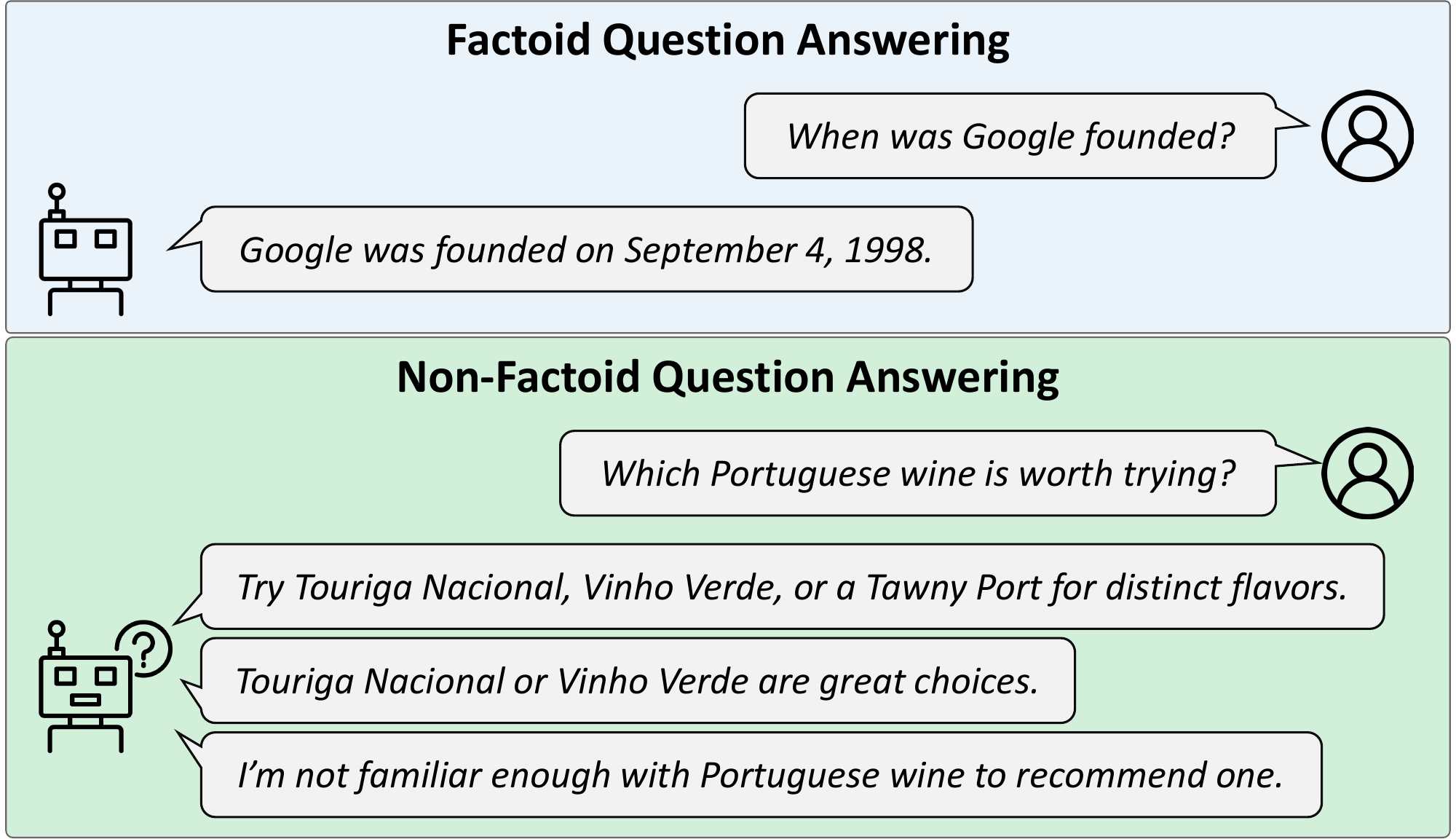}
    \vspace{-5mm}
    \caption{Comparison of factoid question answering (top) and non-factoid question answering (bottom). Factoid questions typically have a single correct answer, whereas non-factoid questions may admit multiple valid answers.}

\vspace{-3mm}
    \label{fig:motivation}
    \vspace{-1mm}
    
\end{figure}

\textbf{Non-factoid} questions (NFQs) represent a fundamentally different challenge from traditional factoid questions, as they require elaborate, interpretive responses that integrate multiple perspectives rather than retrieving single facts \cite{Valeriia2022nfqa-taxonomy}. 
These questions---encompassing comparative evaluations, personal experiences, and open-ended discussions---reflect the rich, multifaceted nature of human information seeking.
Figure~\ref{fig:motivation} illustrates the key differences between factoid questions and NFQs, as well as their example answers.

Despite their prevalence in practical settings \cite{Yang2024Bespoke}, current approaches to non-factoid question answering (NFQA) face significant limitations. 
The core challenge lies in the inherent complexity and heterogeneity of NFQs, which vary along multiple dimensions including question intent, aspect directionality, and degree of contrast between perspectives.

Existing methods fail to adequately address this diversity. 
Type-specific approaches to NFQA \cite{An2024Thread}---methods that focus exclusively on particular NFQ types---struggle to generalize across the full spectrum of NFQ types, whereas retrieval-augmented generation (RAG) systems \cite{Lewis2020RAG, izacard-grave-2021-leveraging}, despite improving contextuality, produce overly homogeneous responses that lack the multi-faceted depth essential for comprehensive answers.
Ultimately, the fundamental challenge is adapting retrieval and generation strategies to the specific characteristics of each NFQ type.

To address these limitations, we propose \method, a type-aware decomposition approach that fundamentally reimagines NFQA within the RAG paradigm. Our approach integrates question type classification directly into the RAG paradigm, enabling tailored retrieval and generation strategies for distinct NFQ types. The key innovation lies in decomposing multi-aspect NFQs into single-aspect sub-queries, allowing targeted retrieval for each aspect before synthesizing a comprehensive response. 
This decomposition approach ensures that generated answers align with user intent while capturing the full complexity of the question.

In order to assess the performance of \method, we evaluate it on Wiki-NFQA, a new benchmark dataset derived from Wikipedia that encompasses a broad spectrum of NFQ types. Our results demonstrate that \method\ significantly outperforms baseline systems, including standard approaches using LLMs and RAG, in handling NFQ complexity and delivering nuanced, intent-aligned answers.

Our main contributions are as follows:

\vspace{-0.075in}

\begin{itemize}[itemsep=0.3mm, parsep=1pt, leftmargin=*]
    \item We propose \textbf{\method}, a novel framework for type-aware decomposition of non-factoid questions that enhances RAG-based NFQA by integrating question type classification with targeted decomposition strategies.
    \item We develop retrieval and generation strategies specifically optimized for different NFQ types, enabling more effective handling of complex and diverse user queries.
    \item We release the \textbf{Wiki-NFQA} dataset, providing a comprehensive benchmark for evaluating QA systems on non-factoid questions and facilitating future NFQA research.
    \item We demonstrate through extensive experiments that \method\ significantly outperforms baseline models, validating the effectiveness of type-aware decomposition in generating contextually appropriate and high-quality answers.
\end{itemize}

By addressing the unique challenges of NFQA through type-aware decomposition, our work bridges the gap between complex user information needs and current QA system capabilities, advancing the development of more adaptive and context-sensitive QA technologies.

\section{Related Work}

\subsection{Non-Factoid Question Answering (NFQA)}

\paragraph{Non-Factoid Question Taxonomy}
To address the complexity of real-world QA, prior work has developed detailed taxonomies for question types \cite{burger2003issues, chaturvedi2014joint, Valeriia2022nfqa-taxonomy}. Unlike factoid questions, which seek concise factual answers, non-factoid questions (NFQs) require subjective, multi-faceted responses \cite{chaturvedi2014joint, Valeriia2022nfqa-taxonomy}. \citet{Valeriia2022nfqa-taxonomy} categorized NFQs into six types: \texttt{Evidence-based}, \texttt{Comparison}, \texttt{Experience}, \texttt{Reason}, \texttt{Instruction}, and \texttt{Debate}. 
Recently, \citet{mishra2025long} emphasized critical challenges faced by LLM-based QA systems when handling NFQs, particularly those requiring in-depth reasoning or nuanced debate, illustrated through practical examples such as detailed descriptive questions (e.g., \textit{``How was the construction of the Taj Mahal perceived by the citizens of Agra in the 17th century?''}). Their work also highlighted specific application contexts where robust NFQA systems are vital, including voice assistants like Amazon Alexa \cite{Hashemi2020ANTIQUE} and web forums frequently hosting user-generated descriptive queries \cite{bajaj2016ms}. 
In contrast to earlier approaches that target individual NFQ types, our method provides a unified framework to handle all categories effectively.

\begin{figure*}[ht!]
    \centering
    \includegraphics[width=\textwidth]{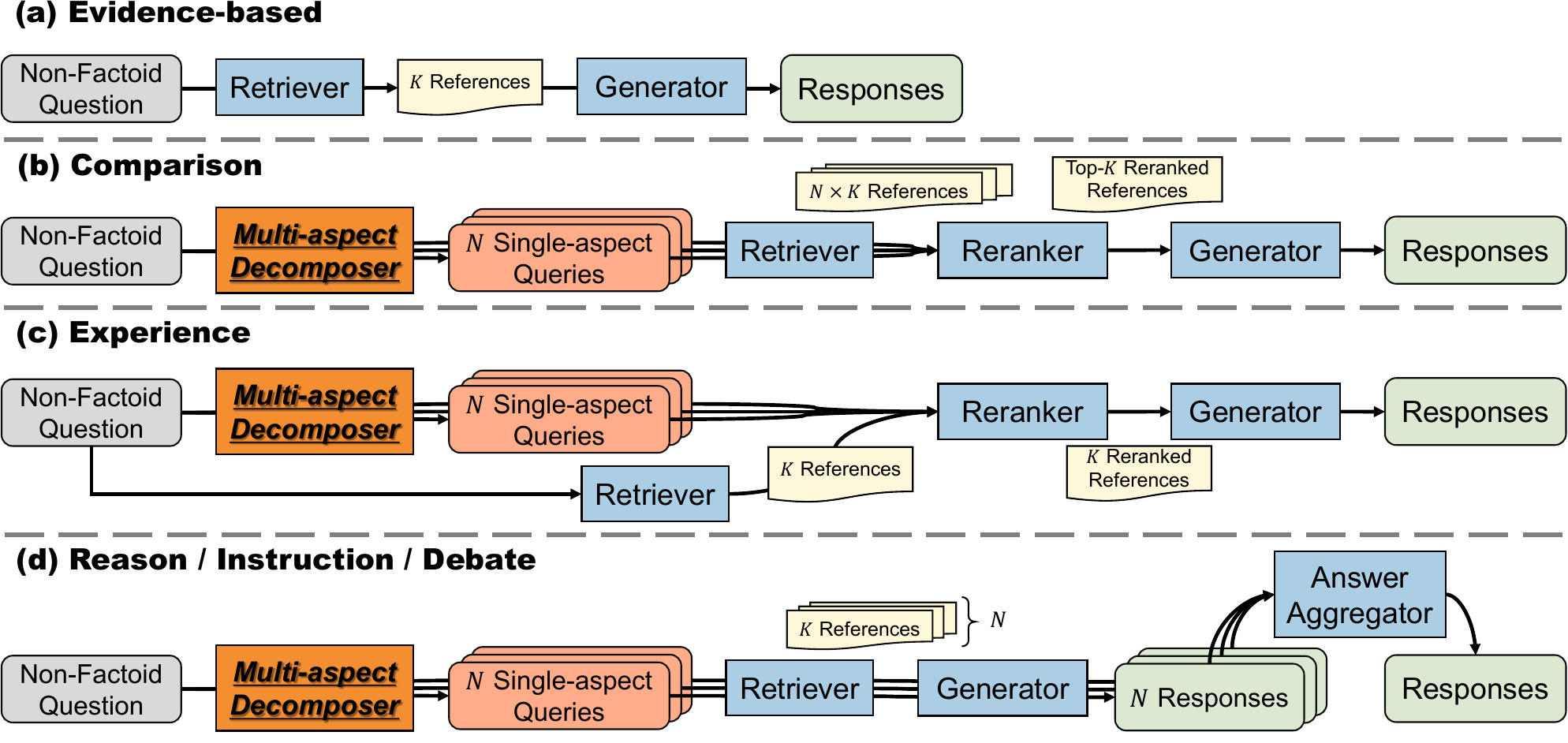}
    \caption{Overview of \method. Non-factoid questions (NFQs) are first classified by a pretrained Type Classifier and then processed according to their type. A Multi-aspect Decomposer and Answer Aggregator address each type’s specific requirements using LLMs with tailored prompts. See Appendix \ref{sec:appendix_method_prompt} and Figure \ref{fig:detailed_overview} for prompt details and a complete illustration.}
    \label{fig:overview}
\end{figure*}


\paragraph{Evaluation Metrics}

Traditional metrics---such as ROUGE or BERTScore \cite{Zhang2020BERTScore}---often fall short in capturing the semantic richness and nuanced quality of NFQA outputs. To overcome these limitations, \citet{yang-etal-2024-linkage} introduced LINKAGE, a listwise ranking framework that uses an LLM as a scorer to rank candidate answers against quality-ordered references. LINKAGE shows stronger correlation with human judgments and outperforms conventional metrics, highlighting its suitability for NFQA evaluation.

\subsection{Retrieval-Augmented Generation (RAG)}

Retrieval-augmented generation (RAG) improves the quality of LLM responses by incorporating external documents, improving factual accuracy and contextual relevance  \cite{Lewis2020RAG, izacard-grave-2021-leveraging}. 
However, the quality of RAG outputs critically depends on the retrieval step: irrelevant or noisy documents can exacerbate hallucinations \cite{Huang2023hallucinationsurvey, Lee2025REFIND}. Recent advances apply query rewriting and multi-hop decomposition to enrich retrieved contexts \cite{rackauckas2024rag-fusion, chan2024rqrag}, and adaptive retrieval strategies that assess query-document relevance before the generation process \cite{jeong2024adaptive, Yan2024Corrective, asai2024selfrag}.

Despite these developments, the application of RAG to NFQA remains underexplored. For example, \citet{Deng2024NFQAGraphMultiHop} proposed a graph-based multi-hop approach for NFQA, but it neither leverages large-scale pretrained LLMs nor employs dynamic retrieval. 
Likewise, \citet{An2024Thread} integrated logic-based threading with RAG for \textit{How-To} questions, yet their method does not generalize across all NFQ types. Our work bridges this gap by combining type-specific decomposition with adaptive retrieval and generation in a single RAG framework.

\begin{table*}[ht!]
\resizebox{\textwidth}{!}{
\begin{tabular}{@{}ll@{}}
\toprule
\textbf{NFQ Type}       & \textbf{Example of Non-Factoid Question}                                                                                                                                                                     \\ \midrule \midrule
\texttt{Evidence-based} & ``How does sterilisation help to keep the money flow even?''                                                                                                                            \\ \midrule
\texttt{Comparison}     & ``what is the difference between dysphagia and odynophagia''                                                                                                                            \\ \midrule
\texttt{Experience}     & ``What are some of the best Portuguese wines?''                                                                                                                                         \\ \midrule
\texttt{Reason}         & \begin{tabular}[c]{@{}l@{}}``Kresy, which roughly was a part of the land beyond the so-called Curson Line, \\ \quad was drawn for what reason?''\end{tabular}                                 \\ \midrule
\texttt{Instruction}    & ``How can you find a lodge to ask to be a member of?''                                                                                                                                  \\ \midrule
\texttt{Debate}         & \begin{tabular}[c]{@{}l@{}}``I Can See Your Voice, a reality show from South Korea, offers what kind of \\ \quad performers a chance to make their dreams of stardom a reality?''\end{tabular} \\ \bottomrule
\end{tabular}
}
\caption{Example non-factoid questions in the Wiki-NFQA dataset, highlighting each NFQ type.}

\label{tab:NFQ_type_examples}
\end{table*}

\section{Method}
In this section, we introduce \method, a novel RAG pipeline designed specifically for non-factoid question (NFQ) types. 
Figure \ref{fig:overview} visually illustrates the overall processing mechanisms of the Multi-aspect Decomposer and the Answer Aggregator for each NFQ type.

NFQs are classified into one of six types using a pre-trained Type Classifier \cite{Valeriia2022nfqa-taxonomy}. Subsequently, queries are preprocessed through the Multi-aspect Decomposer, where each type's characteristics and the underlying intent of the questions are considered. The Multi-aspect Decomposer primarily consists of two modules: the Single-aspect Query Generator and the Keyword Extractor. These modules operate selectively based on the question type and perspective, leveraging few-shot learning and prompt engineering techniques to effectively transform queries according to their respective categories.
The decomposed queries are then processed by one or multiple retrievers to retrieve highly relevant passages. Optionally, the retrieved passages may be re-ranked using a reranker to enhance the quality of results. 

Based on the retrieved information, the generator produces the final answers. If multiple candidate answers are available, the Answer Aggregator integrates these candidates to form a unified response. 
This process is structured based on prompt engineering. Detailed prompt configurations for each module are presented in Appendix \ref{sec:appendix_method_prompt}.

As previously discussed, NFQs inherently involve multiple perspectives, making them challenging to handle effectively using conventional RAG approaches. Considering these distinct characteristics, we propose a type-aware pipeline specifically tailored to NFQs, capable of generating responses that accurately reflect user intent and address the inherent complexity of these questions.

The subsequent subsections elaborate on the definitions of each NFQ type and the corresponding detailed processing strategies.

\subsection{Evidence-based} 
\texttt{Evidence-based} type questions aim to clarify the characteristics or definitions of specific concepts, objects, or events. These questions require precise and reliable factual information. 
These questions inherently have a single aspect, eliminating the need for complex contextual reasoning or multi-aspect decomposition. The intent behind these questions is to obtain clear and concise explanations grounded in evidence, resulting in responses consistently centered around a single aspect.

Accordingly, a straightforward RAG approach is applied to \texttt{Evidence-based} type questions. The retriever utilizes the original question as a query to search relevant documents, and the generator then produces responses based directly on these documents. It is essential to maintain a clear and concise information flow without considering multiple perspectives, thereby ensuring straightforward and accurate answers.

\subsection{Comparison} 
\texttt{Comparison} type questions aim to identify differences, similarities, or superiority among two or more items. 
These questions can have different intentions and must be tailored to the purpose and targets of the comparison.
\texttt{Comparison} type questions can be broadly classified into two categories based on intent: related aspects, focusing on similarities, and contrasting aspects, emphasizing differences or superiority. Consequently, \texttt{Comparison} type questions inherently involve multiple aspects.

Thus, a Multi-aspect Decomposer is required for \texttt{Comparison} type questions. Initially, a Keyword Extractor identifies the purpose of comparison (\textit{compare\_type}) and the items being compared (\textit{keywords\_list}). 
The purpose of the comparison is predefined as one of three types: difference, similarity, or superior. These types are explicitly extracted from the question to determine the scope of the comparison. 
Detailed prompt templates and examples used in this process are described in Appendix \ref{sec:appendix_COMPARISON_prompt}. 

Subsequently, the retriever searches for documents related to each keyword, eliminates redundant results, and reranks the remaining documents based on relevance. Finally, the generator synthesizes the information to produce a balanced response aligned with the comparison purpose. Collecting and integrating information across various comparison criteria and perspectives is crucial for accurately addressing user intent.

\begin{table*}[ht!]
\resizebox{\textwidth}{!}{
    \begin{tabular}{lccccccr}
    \toprule
    \textbf{NFQ Type} & \textbf{NQ-NF} & \textbf{SQD-NF} & \textbf{TQA-NF} & \textbf{2WMHQA-NF} & \textbf{HQA-NF} & \textbf{MSQ-NF} & \textbf{Total} \\ \midrule \midrule
    Evidence-based         & 99             & 130               & 251                  & 10                          & 22                   & 43                  & 555 (58.73\%)  \\
    Comparison             & 5              & 18                & 4                    & 0                           & 8                    & 1                   & 36 (3.81\%)    \\
    Experience             & 0              & 20                & 8                    & 1                           & 10                   & 2                   & 41 (4.34\%)    \\
    Reason                 & 19             & 85                & 23                   & 55                          & 15                   & 21                  & 218 (23.07\%)  \\
    Instruction            & 2              & 21                & 3                    & 8                           & 4                    & 11                  & 49 (5.19\%)    \\
    Debate                 & 1              & 26                & 7                    & 5                           & 3                    & 4                   & 46 (4.87\%)    \\ \midrule
    \textbf{Total}         & 126            & 300               & 296                  & 79                          & 62                   & 82                  & \textbf{945}   \\ \bottomrule
    \end{tabular}
}
\caption{Wiki-NFQA dataset statistics by NFQ type.}
\label{tab:statistics}
\end{table*}

\subsection{Experience} 
\texttt{Experience} type questions seek advice, recommendations, or personal insights, with responses based primarily on individual experiences. These questions naturally involve multiple aspects, and answers can vary significantly due to subjective differences among respondents. Thus, clearly understanding the user's intent and defining key perspectives is essential to provide informative answers encompassing diverse opinions and experiences.

Similar to \texttt{Comparison} type questions, \texttt{Experience} type questions require multi-aspect consideration; however, the focus is not on comparisons based on specific features or criteria but rather on reflecting broader and more comprehensive experiences and diverse opinions. Therefore, \texttt{Experience} type questions require multi-aspect decomposition.

Initially, the Keyword Extractor identifies the primary topics that users seek experiences about and extracts key entities reflecting the question's intent. Specific examples of the prompts used in this process are detailed in Appendix \ref{sec:appendix_EXPERIENCE_prompt}. Following this step, the retriever searches for related documents using these extracted keywords. 
The retrieved documents are then re-ranked according to their similarity to the extracted keywords. Finally, the generator produces an optimized response by synthesizing information that aligns with the user's intent and incorporates diverse perspectives. This ensures the response effectively meets the user's expectations.

\subsection{Reason/Instruction} 
\texttt{Reason} and \texttt{Instruction} type questions both aim to provide information necessary for understanding phenomena or solving problems, but they differ significantly in intent and response approach.

The purpose of \texttt{Reason} type questions is to identify the causes of phenomena or events. These questions require multi-faceted consideration because explanations can vary depending on contextual factors and conditions. 
Different assumptions and conditions may yield multiple possible explanations, resulting in responses often including contrasting or conflicting information.

In contrast, \texttt{Instruction} type questions focus on procedural steps or methodologies. Although the procedures or methods can vary based on specific goals or requirements and thus involve multiple aspects, the responses tend to align similarly rather than diverging significantly, unlike \texttt{Reason} type questions. While various procedures may exist, their fundamental structure or concepts often remain interconnected.

For both \texttt{Reason} and \texttt{Instruction} type questions, a Multi-aspect Decomposer is applied. Initially, a Single-aspect Query Generator decomposes the original query into separate single-aspect queries. 
The retriever and generator then process each query individually, producing separate responses.
An Answer Aggregator subsequently integrates these responses to deliver a clear, systematically organized final answer. Examples of the prompts used in this process are detailed in Appendix \ref{sec:appendix_REASON_INSTRUCTION_prompt}.

\subsection{Debate}
\texttt{Debate} type questions focus on controversial topics and aim to explore and reflect multiple perspectives, especially opposing perspectives. These questions inherently possess multiple perspectives and require the inclusion of contrasting arguments and perspectives, as subjective positions may vary according to underlying assumptions and perspectives, unlike factual questions.

\begin{figure*}[th!]
    \centering
    \begin{tcolorbox}[colback=gray!10, colframe=black, title=Prompt Template for LINKAGE]
        Please impartially rank the given candidate answer to a non-factoid question accurately within the reference answer list, which are ranked in descending order of quality. The top answers are of the highest quality, while those at the bottom may be poor or unrelated.\\
        Determine the ranking of the given candidate answer within the provided reference answer list. For instance, if it outperforms all references, output [[1]]. If it's deemed inferior to all four references, output [[4]].\\
        Your response must strictly following this format: "[[2]]" if candidate answer could rank 2nd.\\
        Below are the user's question, reference answer list, and the candidate answer.\\
        Question:\{\texttt{question}\}\\
        Reference answer list:\{\texttt{reference\_answers}\}\\
        Candidate answer:\{\texttt{candidate\_answer}\}
    \end{tcolorbox}
    \vspace{-3mm}
    \caption{The LINKAGE prompt template from \citet{yang-etal-2024-linkage} used to rank candidate answers in our evaluation of \method\ and the baselines.}
    \label{fig:LINKAGE_prompt}
\end{figure*}

To effectively respond to \texttt{Debate} type questions, it is essential to fairly represent the logic of each opposing perspective and generate unbiased, balanced responses. Thus, a Multi-aspect Decomposer breaks down the question into debate topics and diverse opinions. The Single-aspect Query Generator then formulates individual queries for each opinion. 
The retriever and generator process each query separately, producing individual responses.
Finally, an LLM with a debate mediator persona \cite{liang-etal-2024-encouraging} synthesizes these diverse perspectives, generating a balanced final response from a mediator's perspective. Detailed prompts applied in the Single-aspect Query Generator and Debate Mediator processes are outlined in Appendix \ref{sec:appendix_DEBATE_prompt}. This approach ensures responses fairly and transparently reflect diverse perspectives, enabling comprehensive and balanced information delivery suitable for \texttt{Debate} type questions.

\begin{table*}[ht]
\resizebox{\textwidth}{!}{

\begin{tabular}{lllcccccc}
\toprule
\multirow{2}{*}{\textbf{Model}}                & \multirow{2}{*}{\textbf{Scorer LLM}}              & \multirow{2}{*}{\textbf{Methods}} & \multicolumn{6}{c}{\textbf{Wiki-NFQA Dataset}} \\
\cmidrule(lr){4-9}
                              &                              &                  & \textbf{NQ-NF}  & \textbf{SQD-NF} & \textbf{TQA-NF}  & \textbf{2WMH-NF} & \textbf{HQA-NF} & \textbf{MSQ-NF} \\
\midrule
\midrule
\multirow{6}{*}{Llama-3.2-3B} & \multirow{3}{*}{Mistral-7B}  & LLM              & 0.5893          & 0.5119          & 0.6191            & 0.3565          & 0.4825            & 0.4262           \\
                              &                              & RAG              & 0.5294          & 0.4944          & 0.5470            & 0.4150          & 0.4530            & 0.4047           \\
                              &                              & \method          & \textbf{0.7659} & \textbf{0.6493} & \textbf{0.7061}   & \textbf{0.4544} & \textbf{0.5624}   & \textbf{0.5356}  \\
\cmidrule(lr){2-9}
                              & \multirow{3}{*}{GPT-4o mini} & LLM              & 0.4934          & 0.4506          & 0.5380            & 0.3070          & 0.3669            & 0.2917           \\
                              &                              & RAG              & 0.4187          & 0.3553          & 0.4586            & 0.2859          & 0.2957            & 0.2866           \\
                              &                              & \method          & \textbf{0.8366} & \textbf{0.7139} & \textbf{0.7013}   & \textbf{0.3692} & \textbf{0.5470}   & \textbf{0.4482}           \\
\midrule
\midrule
\multirow{6}{*}{Mistral-7B}   & \multirow{3}{*}{Mistral-7B}  & LLM              & 0.6356          & 0.5450          & 0.6363            & \textbf{0.4821} & 0.5255            & \textbf{0.5081}  \\
                              &                              & RAG              & 0.5635          & 0.5069          & 0.6233            & 0.4789          & 0.5323            & 0.4438           \\
                              &                              & \method          & \textbf{0.7103} & \textbf{0.6333} & \textbf{0.6709}   & 0.4747          & \textbf{0.6035}   & 0.4512           \\ 
\cmidrule(lr){2-9}
                              & \multirow{3}{*}{GPT-4o mini} & LLM              & 0.4656          & 0.4222          & 0.5921            & 0.3175          & 0.3965            & 0.3384           \\
                              &                              & RAG              & 0.4411          & 0.3817          & 0.5450            & 0.2890          & 0.3562            & 0.3079           \\
                              &                              & \method          & \textbf{0.8413} & \textbf{0.7444} & \textbf{0.7767}   & \textbf{0.3987} & \textbf{0.6653}   & \textbf{0.4929}  \\ 
\bottomrule
\end{tabular}

}

\caption{Evaluation on the Wiki-NFQA dataset comparing various language models, scorer LLMs, and methods using Mean Reciprocal Rank (MRR). Answers were ranked with LINKAGE \cite{yang-etal-2024-linkage} and evaluated by MRR.}
\label{tab:MRR_results}
\end{table*}

\section{Experimental Setup}

\subsection{Model}
We compare our \method\ with \textbf{LLM-} and \textbf{RAG-based} QA systems as baselines.
In our experiments, we use a black-box LLM and two open-weight LLMs with different numbers of parameters: (i) \href{https://huggingface.co/meta-llama/Llama-3.2-3B-Instruct}{Llama-3.2-3B-Instruct} (Llama-3.2-3B), (ii) \href{https://huggingface.co/mistralai/Mistral-7B-Instruct-v0.2}{Mistral-7B-Instruct-v0.2} \citep[Mistral-7B;][]{Jiang2023Mistral-7B}, and (iii) GPT-4o-mini-2024-07-18 (GPT-4o mini).

All inputs to the LLMs (including the RAG generator) are formatted using prompt templates. 
The prompt templates used in our experiments are provided in Appendix \ref{sec:appendix_baseline_prompt}.

\subsection{Listwise Ranking Evaluation (LINKAGE)}
To evaluate non-factoid question answering (NFQA) systems, we adopt LINKAGE \cite{yang-etal-2024-linkage}, a listwise ranking framework designed for NFQA. LINKAGE orders each candidate answer by comparing its quality against a reference list of answers, defined formally as:
\begin{align}
  \label{eq:LINKAGE}
  R_{i} &= \{ r_{i_{1}}, r_{i_{2}}, \dots, r_{i_{n}} \}, \\
  rank_{c_{i}} &= \mathrm{LLM}\bigl(\mathcal{P}_{L}, q_{i}, c_{i}, R_{i}\bigr).
\end{align}
Here, $q_{i}$ is the $i$-th question, $c_{i}$ is the candidate answer under evaluation, and $R_{i}$ is the set of reference answers $\{r_{i_k}\}$ sorted from highest to lowest quality. The scorer LLM, guided by the LINKAGE prompt $\mathcal{P}_{L}$, assigns $rank_{c_{i}}$ based on each candidate’s position within $R_{i}$. 
The complete LINKAGE prompt template is shown in Figure~\ref{fig:LINKAGE_prompt}.

\paragraph{Ranking Metrics}  

To quantify LINKAGE rankings, we employ two complementary metrics: Mean Reciprocal Rank (MRR) \cite{voorhees-tice-2000-trec} and Mean Percentile Rank (MPR):
\begin{align}
  \label{eq:ranking_metrics}
  \mathrm{MRR} &= \frac{1}{N} \sum_{i=1}^{N} \frac{1}{rank_{c_{i}}}, \\
  \mathrm{MPR} &= \frac{1}{N} \sum_{i=1}^{N} \biggl(1 - \frac{rank_{c_{i}} - 1}{\lvert R_{i}\rvert}\biggr) \times 100.
\end{align}
MRR measures how close candidate answers rank to the top of the list, with higher values indicating better performance. MPR converts each rank into a percentile, reflecting the candidate’s relative position within its reference list; higher MPR scores denote superior overall ranking across all positions. Together, MRR highlights top-answer accuracy, while MPR provides insight into performance across the entire ranking spectrum.

\subsection{Dataset Construction}

To test the NFQA methods, we curate the \textbf{Wiki-NFQA dataset}, a specialized resource tailored for NFQA. It is derived from existing Wikipedia-based datasets:  Natural Questions \citep[NQ;][]{kwiatkowski-etal-2019-natural}, SQuAD \citep[SQD;][]{rajpurkar-etal-2016-squad}, TriviaQA \citep[TQA;][]{joshi-etal-2017-triviaqa}, 2WikiMultiHopQA \citep[2WMH;][]{ho-etal-2020-constructing}, HotpotQA \citep[HQA;][]{yang-etal-2018-hotpotqa}, MuSiQue \citep[MSQ;][]{trivedi2022musique}. 

\paragraph{Filtering Non-Factoid Questions}
Through a systematic filtering process, we extract non-factoid questions, then generate high-quality reference answers to ensure the dataset's suitability for NFQA evaluation.

We use the nf-cats\footnote{\url{https://huggingface.co/Lurunchik/nf-cats}} \cite{Valeriia2022nfqa-taxonomy}, a RoBERTa-based pre-trained NFQ category classifier, to extract NFQs from existing Wikipedia-based datasets.
Since it categorizes questions into factoid and non-factoid types, we only retain  those classified as non-factoid for further processing.
To ensure a more rigorously curated dataset, we filter the data using heuristics based on the question patterns outlined in the NFQ taxonomy proposed by \citet{Valeriia2022nfqa-taxonomy}.
Table \ref{tab:statistics} presents the statistics for the Wiki-NFQA dataset.

\paragraph{Reference Answers Generation}
Since these datasets only have a single-grade ground truth answer, we generate diverse reference answers of varying quality for LINKAGE evaluation, as described in \citet{yang-etal-2024-linkage}. 
After constructing the reference answers, we use the GPT-4o-2024-11-20 to annotate their quality level.
Prompt details about generating reference answers are provided in Appendix \ref{sec:appendix_reference_construction}, and \ref{sec:appendix_annotation}. 

\begin{figure*}[ht!]
  \centering
  \includegraphics[width=\textwidth]{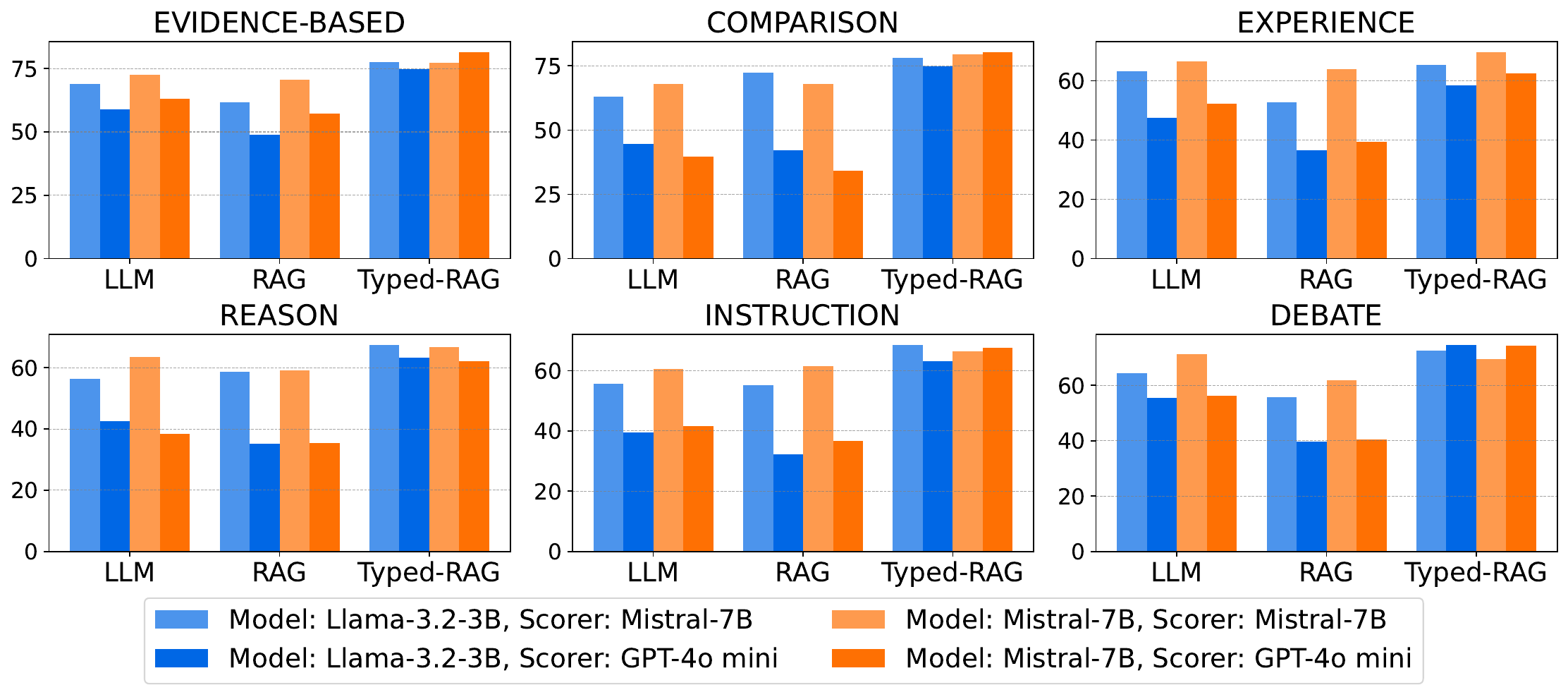}
  \caption{Comparison of Mean Percentile Rank (MPR) for LLMs, RAGs, and \method\ across six NFQ categories in the Wiki-NFQA dataset. Results are shown for two model setups (Llama-3.2-3B and Mistral-7B) and two scorer LLMs (Mistral-7B and GPT-4o mini); the y-axis displays MPR (\%), where higher values indicate better performance.}

  \label{fig:MPR_results}
\end{figure*}


\paragraph{Examples of the Wiki-NFQA Dataset}  

Table~\ref{tab:NFQ_type_examples} provides examples of questions that represent each type of NFQ in the Wiki-NFQA dataset. 

The \texttt{Evidence-based} type questions require answers grounded in verifiable sources, while \texttt{Comparison} type questions seek distinctions between concepts. 
\texttt{Experience} type questions solicit subjective opinions or recommendations, while \texttt{Reason} type questions aim to uncover the rationale behind events or concepts. 
\texttt{Instruction} type questions request procedural guidance, and \texttt{Debate} type questions involve discussions on controversial or interpretive topics.

\section{Experimental Results}

We evaluate \method\ on the Wiki-NFQA dataset across all NFQ categories and model configurations. As shown in Table \ref{tab:MRR_results} (MRR) and Figure \ref{fig:MPR_results} (MPR), \method\ consistently outperforms both LLM- and RAG-based baselines. These metrics demonstrate that our approach not only elevates the ranking of generated answers but also improves their relative quality. Scorer LLMs uniformly rate \method’s responses as more relevant and comprehensive. Representative examples of \method’s outputs for each non-factoid question type are provided in Appendix \ref{sec:appendix_case_study}.

\subsection{Impact of Scorer LLMs and Base Models}

The performance of all methods depends on the choice of scorer LLM and base model. 
Since each LLM evaluates responses using its own learned criteria and internal representations, scores from different scorers should not be compared directly.
Instead, performance comparisons are most meaningful when considering the relative ranking of methods under the same scorer.

As reported in Table \ref{tab:MRR_results} and Figure \ref{fig:MPR_results}, scores generally decrease when switching from Mistral-7B to GPT-4o mini as the scorer. This trend likely stems from GPT-4o mini’s greater sophistication and stricter evaluation standards, which penalize even minor inconsistencies or lack of depth. This pattern holds across all base models and methods, underscoring that more powerful scorers apply more stringent criteria. Notably, despite the overall score reductions, \method\ retains a clear advantage over both LLM- and RAG-based baselines, demonstrating its robustness to changes in scorer strictness.

\subsection{Limitations of RAG and Benefits of \method}

Our experiments also reveal that RAG-based methods underperform direct LLM-based generation (see Table \ref{tab:MRR_results} and Figure \ref{fig:MPR_results}). We attribute this shortfall to the noise introduced by retrieved factual information, which can hinder response generation in NFQA tasks. \method\ addresses this challenge through a multi-aspect decomposition strategy that structures retrieval around the distinct facets of non-factoid questions. By reducing irrelevant noise and ensuring more focused retrieval, \method\ consistently outperforms both RAG and LLM-only approaches—particularly on reasoning-intensive subsets—thereby enhancing the overall quality of generated answers.

\section{Conclusion}

In this paper, we introduced \method, a novel RAG-based framework for non-factoid question answering (NFQA) that incorporates type-aware multi-aspect decomposition. By first classifying each NFQ into a specific category and then decomposing it into focused sub-queries, \method\ enables targeted retrieval and answer generation for each aspect. The retrieved sub-responses are aggregated to produce comprehensive, nuanced answers that better address the diverse requirements of non-factoid questions.

To support evaluation, we also curated Wiki-NFQA, a benchmark dataset covering a wide range of NFQ types. 
Experimental results on Wiki-NFQA dataset show that \method\ consistently outperforms both LLM-only and standard RAG baselines across all question types and scorer LLM settings. These findings validate the effectiveness and robustness of type-aware multi-aspect decomposition in enhancing both retrieval quality and answer relevance for NFQA.

Future work could explore extending \method\ to incorporate more fine-grained question types and further refine the decomposition strategies.
Additionally, applying our approach to other specific domains or applying fine-tuning methods and integrating it with more sophisticated retrieval mechanisms could further improve the performance and adaptability of NFQA systems.
\section*{Limitations}
Our work is the first to introduce RAG to NFQA, but it has several limitations.

A key limitation is the absence of a direct comparison between \method\ and existing query rewriting and decomposition methodologies. 
Although \method\ provides a structured approach to these tasks, its performance relative to other techniques remains unexplored.
Though various query rewriting and decomposition techniques have been proposed to improve retrieval quality, this study does not empirically evaluate the effectiveness of query reformulation, retrieval relevance, or computational overhead of \method\ relative to these approaches. 
A systematic comparison with these methods would provide a clearer understanding of \method's advantages and limitations. Future work should incorporate benchmark evaluations against these established techniques to better position \method\ within the landscape of query rewriting and decomposition research.

Another limitation of our evaluation setup is that we use the same model to assess the quality of its generated responses. This self-evaluation approach may introduce bias because the model may struggle to distinguish differences in quality among the answers it produced. To mitigate this issue, future work could explore using stronger LLMs, human assessments, or ensemble scoring methods for evaluation. Adopting these strategies would improve the reliability of quality assessments and reduce potential biases in our evaluation framework.
\section*{Acknowledgments}

This research was supported by Brian Impact Foundation, a non-profit organization dedicated to the advancement of science and technology for all.
A prior version of this work was presented in a non-archival form at the NAACL 2025 Student Research Workshop.

\bibliography{custom}

\appendix
\clearpage
\onecolumn

\section{Prompt Details}
\label{sec:appendix_prompt_details}

\subsection{Reference List Construction}
\label{sec:appendix_reference_construction}

\begin{figure*}[htb!]
    \centering
    \begin{tcolorbox}[colback=gray!10, colframe=black, title=Prompt Template to Generate the Highest Standard Reference Answer]
        Given a non-factoid question:"\{\texttt{question}\}" and its answer:"\{\texttt{ground\_truth}\}"\\
        Use your internal knowledge to rewrite this answer.
    \end{tcolorbox}
    \vspace{-3mm}
    \caption{Prompt template proposed by \citet{yang-etal-2024-linkage} to generate the highest standard reference answer using LLM's internal knowledge.}
    \label{fig:Prompt_to_generate_std_answer}
\end{figure*}

\begin{figure*}[htb!]
    \small
    \centering
    \begin{tcolorbox}[colback=gray!10, colframe=black, title=Prompt Template to Generate Diverse Qualities of Reference Answers]
        Generate three different answers to a non-factoid question from good to bad in quality, each inferior to the golden answer I give you. 
        Ensure that the quality gap from good to bad is very significant among these three answers. 
        Golden answer is the reasonable and convincing answer to the question. 
        Answer 1 can be an answer to the question, however, it is not sufficiently convincing. 
        Answer 2 does not answer the question or if it does, it provides an unreasonable answer. 
        Answer 3 is completely out of context or does not make any sense.
        \\
        \\
        Here are 3 examples for your reference.
        \\
        1.Non-factoid Question: how can we get concentration on something?
        \\
        Golden Answer: To improve concentration, set clear goals, create a distraction-free environment, use time management techniques like the Pomodoro Technique, practice mindfulness, take regular breaks, stay organized, limit multitasking, practice deep work, maintain physical health, and seek help if needed.
        \\
        Output:
        \\
        Answer 1: Improve focus: set goals, quiet space, Pomodoro Technique, mindfulness, breaks, organization, limit multitasking, deep work, health, seek help if needed.
        \\
        Answer 2: Just like and enjoy the work you do, concentration will come automatically.
        \\
        Answer 3: If you are student, you should concentrate on studies and don't ask childish questions.
        \\
        \\
        2.Non-factoid Question: Why doesn't the water fall off earth if it's round?
        \\
        Golden Answer: Earth's gravity pulls everything toward its center, including water. Even though Earth is round, gravity keeps water and everything else anchored to its surface. Gravity's force is strong enough to counteract the Earth's curvature, preventing water from falling off.
        \\
        Output:
        \\
        Answer 1: This goes along with the question of why don't we fall off the earth if it is round. The answer is because gravity is holding us (and the water) down.
        \\
        Answer 2: Same reason the people don't.
        \\
        Answer 3: When rain drops fall through the atmosphere CO2 becomes dissolved in the water. CO2 is a normal component of the Earth's atmosphere, thus the rain is considered naturally acidic.
        \\
        \\
        3.Non-factoid Question: How do I determine the charge of the iron in FeCl3?
        \\
        Golden Answer: Since chloride ions (Cl-) each carry a charge of -1, and there are three chloride ions in FeCl3, the total negative charge from chloride ions is -3. To balance this, the iron ion (Fe) must have a charge of +3 to ensure the compound has a neutral overall charge. Therefore, the charge of the iron ion in FeCl3 is +3.
        \\
        Output:
        \\
        Answer 1: Charge of Fe in Fecl3 is 3. Iron has either 2 as valancy or 3. in this case it bonds with three chlorine molecules. therefore its valency and charge is three.
        \\
        Answer 2: If two particles (or ions, or whatever) have opposite charge, then one has positive charge and one has negative charge.
        \\
        Answer 3: take a piece of iron. Wrap a copper wire around the iron in tight close coils. run a charge through the wire.
        \\
        \\
        Below are the non-factoid question, and the golden answer.
        \\
        Non-factoid Question: \{\texttt{question}\}
        \\
        Golden Answer: \{\texttt{ground\_truth}\}
        \\
        Output:
    \end{tcolorbox}
    \vspace{-3mm}
    \caption{Prompt template proposed by \citet{yang-etal-2024-linkage} to generate diverse qualities of reference answers.}
    \label{fig:Prompt_to_generate_diverse_qualities}
\end{figure*}

\clearpage

\subsection{Reference Answers Annotation}
\label{sec:appendix_annotation}

\begin{figure*}[htb!]
    \centering
    \begin{tcolorbox}[colback=gray!10, colframe=black, title=System Prompt for Reference Answers Annotation]
        Your task is to evaluate the relevance and quality of multiple candidate answers for a given non-factoid question.\\
        Please evaluate the quality of each answer in a step-by-step manner.\\
        Follow the structured guidelines below to ensure consistency and accuracy in your evaluation.\\
        \\
        \# Notes on Candidate Answers\\
        Multiple candidate answers can come in two forms:\\
        - Single choice answer: A single string, e.g., \`{}"born again"\`{}.\\
        - Multiple choice answer: A list of strings, e.g., \`{}[`traffic calming', `aesthetics']\`{}.\\
        When evaluating multiple choice answers, treat the entire list as a single unit. Do **not** split them into individual components; instead, evaluate the overall quality as a whole.\\
        \\
        \# Evaluation Criteria\\
        Assign a label to each candidate answer based on the following criteria:\\
        - 3: The answer provides a comprehensive, accurate, and contextually relevant response that directly addresses the question.\\
        - 2: The answer is accurate and relevant but lacks depth or comprehensive coverage.\\
        - 1: The answer is somewhat relevant but contains inaccuracies, vagueness, or insufficient detail.\\
        - 0: The answer is irrelevant, incorrect, or fails to address the question meaningfully.\\
        **If there are two or more answers that you think are close in quality, you can give the same label.**\\
        \\
        \# Response Format  \\
        - Assign a label to each answer strictly in the format: \`{}Answer X: [[Y]]\`{}, where \`{}X\`{} is the answer number, and \`{}Y\`{} is the integer score (0-3).\\
        - Do **not** include any additional comments or explanations outside this format.
    \end{tcolorbox}
    \vspace{-3mm}
    \begin{tcolorbox}[colback=gray!10, colframe=black, title=Input Prompt Template for Reference Answers Annotation]
        \# Inputs  \\
        - Non-Factoid Question: \{\texttt{question}\}  \\
        - Candidate Answers:  \\
        \{\texttt{reference\_answers}\}
    \end{tcolorbox}
    \vspace{-3mm}
    \caption{System prompt (top) and input prompt template (bottom) adapted from \citet{yang-etal-2024-linkage} for annotating the quality level of generated reference answers.}
    \label{fig:Annotation_Prompt}
\end{figure*}

\clearpage

\subsection{Prompt templates for Baseline Methods}
\label{sec:appendix_baseline_prompt}

\begin{figure*}[htb!]
    \centering
    \begin{tcolorbox}[colback=gray!10, colframe=black, title=Prompt template for LLM]
        You are an assistant for answering questions.\\
        Answer the following question.
        \\ \\
        \#\#\# Question\\
        \{\texttt{question}\} \\ \\
        \#\#\# Answer
    \end{tcolorbox}
    \vspace{-3mm}
    \caption{Prompt template for LLM method.}
    \label{fig:LLM_Prompt}
\end{figure*}

\begin{figure*}[htb!]
    \centering
    \begin{tcolorbox}[colback=gray!10, colframe=black, title=Prompt template for RAG]
        You are an assistant for answering questions.\\
        Refer to the references below and answer the following question.
        \\ \\
        \#\#\# References\\
        \{\texttt{reference\_passages}\} \\ \\
        \#\#\# Question\\
        \{\texttt{question}\} \\ \\
        \#\#\# Answer
    \end{tcolorbox}
    \vspace{-3mm}
    
    \caption{Prompt template for RAG method.}
    \label{fig:RAG_Prompt}
\end{figure*}

\clearpage

\subsection{Prompt templates for \method}
\label{sec:appendix_method_prompt}

\subsubsection{Comparison}
\label{sec:appendix_COMPARISON_prompt}

\begin{figure*}[hbt!]
    \centering
    
    \begin{tcolorbox}[colback=gray!10, colframe=black, title=Prompt Template for Keyword Extraction in Comparison Type Questions]
        You are a query analysis assistant. Based on the query type, apply the relevant prompt to transform the query to better align with the user's intent, ensuring clarity and precision.\\
        Determine if the input query is a compare-type question (i.e., compare/contrast two or more things, understand their differences/similarities.) as a "Query Analyst". 
        If so, perform the following:\\
        \\
        1. Identify the type of comparison: "differences", "similarities", or "superiority".\\
        2. Extract the subjects of comparison and represent them as specific, contextualized phrases.\\
        \\
        \#\#\# Output format\\
        \{"is\_compare": true/false, "compare\_type": "", "keywords\_list": []\}\\
        \\
        \#\#\# Example\\
        Query: "Who is more intelligent than humans on earth?"\\
        Analysis:\\
        \{"is\_compare": true, "compare\_type": "superiority", "keywords\_list": ["human intelligence", "the intelligence of other beings"]\}\\
        \\
        \#\#\# Input\\
        Query: \{\texttt{query}\}\\
        \#\#\# Output\\
        Analysis:\\
    \end{tcolorbox}
    \vspace{-3mm}

    \caption{Prompt template for keyword extraction in Comparison type questions.}
    \label{fig:COMPARISON_keyword_extractor_Prompt}
\end{figure*}

\begin{figure*}[hbt!]
    \centering
    
    \begin{tcolorbox}[colback=gray!10, colframe=black, title=Prompt Template for Generating a Response to Comparison Type Questions]
        You are an assistant for answering questions.\\
        You are given the extracted parts of a long document and a question. 
        Refer to the references below and answer the following question.\\\\
        The question is a compare-type with a specific comparison type and keywords indicating the items to compare.\\
        Answer based on this comparison type and the target keywords provided.\\
        \\
        \#\#\# Inputs\\
        Question: \{question\}\\
        Comparison Type: \{\texttt{comparison\_type}\}\\
        Keywords: \{\texttt{keywords}\}\\
        References:\\
        \{\texttt{reference\_passages}\}\\
        \#\#\# Output\\
        Answer: 
    \end{tcolorbox}
    \vspace{-3mm}

    \caption{Prompt template for generating a response to Comparison type questions.}
    \label{fig:COMPARISON_sub_queries_generator_Prompt}
\end{figure*}

\clearpage

\subsubsection{Experience}
\label{sec:appendix_EXPERIENCE_prompt}

Figure \ref{fig:EXPERIENCE_keyword_extractor_Prompt} shows the prompt template for responding to Experience type questions. 
The retrieved passages are subsequently re-ranked based on the extracted keywords.
After reranking, we use the prompt template for RAG (Figure \ref{fig:RAG_Prompt}) to generate answers.

\begin{figure*}[hbt!]
    \centering
    
    \begin{tcolorbox}[colback=gray!10, colframe=black, title=Prompt Template for Keyword Extraction in Experience Type Questions]
        You are a query analysis assistant. Based on the query type, apply the relevant prompt to transform the query to better align with the user's intent, ensuring clarity and precision.\\
        The input question is an experience-type question (i.e., get advice or recommendations on a particular topic.). 
        As a "Query Analyst", please evaluate this question and proceed with the following steps.\\
        \\
        1. Identify the topic intended to be gathered from experience-based questions.\\
        2. Extract the key entities in the question, considering the intent of asking about experience, to facilitate an accurate response.\\
        \\
        \#\#\# Output format\\
        \`{}["Keyword 1", ..., "Keyword N"]\`{} (List of string, separated with comma)\\
        \\
        \#\#\# Example\\
        Question (Input): "What are some of the best Portuguese wines?"\\
        Answer (Output): ["Portuguese wines", "best"]\\
        \\
        \#\#\# Input\\
        Question: \{\texttt{question}\}\\
        \#\#\# Output\\
        Answer: 
    \end{tcolorbox}
    \vspace{-3mm}

    \caption{Prompt template for keyword extraction in Experience type questions.}
    \label{fig:EXPERIENCE_keyword_extractor_Prompt}
\end{figure*}

\clearpage

\subsubsection{Reason \& Instruction}
\label{sec:appendix_REASON_INSTRUCTION_prompt}

\begin{figure*}[hbt!]
    \centering
    \small
    \begin{tcolorbox}[colback=gray!10, colframe=black, title=Prompt Template for Generating Sub-queries in Reason Type Questions]
        You are a query analysis assistant. 
        Based on the query type, apply the relevant prompt to transform the query to better align with the user's intent, ensuring clarity and precision.\\
        The input query is a reason-type question (i.e., a question posed to understand the reason behind a particular concept or phenomenon). 
        As a "Query Analyst", please evaluate this query and proceed with the following steps.\\
        \\
        1. Break down the original instruction into multiple sub-queries that preserve the core intent but use varied language and structure. These multiple sub-queries should aim to capture different linguistic expressions of the original instruction while still aligning with its intended meaning.\\
        2. Create at least 2 to 5 distinct multiple sub-queries.\\
        \\
        \#\#\# Output format\\
        \`{}["sub-query 1", ..., "sub-query N"]\`{} (List of string, separated with comma)\\
        \\
        \#\#\# Input\\
        Query: \{\texttt{query}\}\\
        \#\#\# Output\\
        Multiple sub-queries: 
    \end{tcolorbox}
    \vspace{-3mm}

    \caption{Prompt template for generating sub-queries in Reason type questions.}
    \label{fig:REASON_sub_queries_generator_Prompt}
\end{figure*}

\begin{figure*}[hbt!]
    \centering
    \small
    
    \begin{tcolorbox}[colback=gray!10, colframe=black, title=Prompt Template for Generating Sub-queries in Instruction Type Questions]
        You are a query analysis assistant. 
        Based on the query type, apply the relevant prompt to transform the query to better align with the user's intent, ensuring clarity and precision.\\
        The input query is an instruction-type question (i.e., Instructions/guidelines provided in a step-by-step manner). \\
        As a "Query Analyst", please evaluate this query and proceed with the following steps.\\
        1. Break down the original instruction into multiple sub-queries that preserve the core intent but use varied language and structure.\\
        These multiple sub-queries should aim to capture different linguistic expressions of the original instruction while still aligning with its intended meaning.\\
        2. Create at least 2 to 5 distinct multiple sub-queries.\\
        \\
        \#\#\# Output format\\
        \`{}["sub-query 1", ..., "sub-query N"]\`{} (List of string, separated with comma)\\
        \\
        \#\#\# Input\\
        Query: \{\texttt{query}\}\\
        \#\#\# Output\\
        Multiple sub-queries: 
    \end{tcolorbox}
    \vspace{-3mm}

    \caption{Prompt template for generating sub-queries in Instruction type questions.}
    \label{fig:INSTRUCTION_sub_queries_generator_Prompt}
\end{figure*}

\begin{figure*}[hbt!]
    \centering
    \small
    
    \begin{tcolorbox}[colback=gray!10, colframe=black, title=Prompt Template for Aggregating Answers to an Original Question]
        You are an assistant tasked with aggregating answers to a question.\\
        You are provided with the original question and multiple question-answer pairs. 
        These queries preserve the core intent of the original question but use varied language and structure. 
        Your goal is to review the question-answer pairs and synthesize a concise and accurate response to the original question based on the information provided.
        \\
        Using the information from the question-answer pairs, generate a brief and clear answer to the original question.\\
        \\
        \#\#\# Inputs\\
        Original Question: \{\texttt{original\_question}\}\\
        Question-Answer Pairs: \\
        \{\texttt{qa\_pairs\_text}\}\\
        \#\#\# Output\\
        Aggregated Answer: 
    \end{tcolorbox}
    \vspace{-3mm}

    \caption{Prompt template for aggregating answers to an original question. Used by Reason type questions and Instruction type questions.}
    \label{fig:REASON_INSTRUCTION_aggregator_Prompt}
\end{figure*}

\clearpage

\subsubsection{Debate}
\label{sec:appendix_DEBATE_prompt}

\begin{figure*}[htb!]
    \centering
    \footnotesize
    
    \begin{tcolorbox}[colback=gray!10, colframe=black, title=Prompt Template for Generating Sub-queries in Debate Type Questions]
        \vspace{-1mm}
        You are a query analysis assistant. Based on the query type, apply the relevant prompt to transform the query to better align with the user's intent, ensuring clarity and precision. \\
        The input question is a debate-type question (i.e., invites multiple perspectives). 
        As a "Query Analyst", please evaluate this question and proceed with the following steps.\\
        \\
        1. Extract the debate topic.\\
        2. Identify 2 to 5 key perspectives on this topic.\\
        3. Generate a sub-query reflecting each perspective’s bias.\\
        \\
        Ensure each sub-query fits a Retrieval-Augmented Generation (RAG) framework, seeking passages that align with the viewpoint.\\
        \\
        \#\#\# Output format\\
        \{"debate\_topic": \{topic\},\ "dist\_opinion": [list of perspectives],\ "sub-queries": \{"opinion1": "biased sub-query for opinion1",\ "opinion2": "biased sub-query for opinion2",\ ...\}\\
        \\
        \#\#\# Example\\
        Query: "Is Trump a good president?"\\
        Answer: \\
        \{\\
        \phantom{xxxx}"debate\_topic": "Donald Trump's presidency",\\
        \phantom{xxxx}"dist\_opinion": ["positive", "negative", "neutral"],\\
        \phantom{xxxx}"sub-queries": \{\\
        \phantom{xxxx}\phantom{xxxx}"positive": "Was Donald Trump one of the best presidents for economic growth?",\\
        \phantom{xxxx}\phantom{xxxx}"negative": "Did Trump's presidency harm the U.S. economy and leadership?",\\
        \phantom{xxxx}\phantom{xxxx}"neutral": "Can we assess Trump's tenure's strengths and weaknesses?"\\
        \phantom{xxxx}\}\\
        \}\\
        \\
        \#\#\# Input\\
        Query: \{\texttt{query}\}\\
        \#\#\# Output\\
        Answer: 
        \vspace{-1mm}
        
    \end{tcolorbox}
    
    \begin{tcolorbox}[colback=gray!10, colframe=black, title=Prompt Template for Debate Mediator in Debate Type Questions]
        \vspace{-1mm}
        You are acting as the mediator in a debate.\\
        Below is a topic and responses provided by n participants, each with their own perspective. 
        Your task is to synthesize these responses by considering both the debate topic and each participant's viewpoint, providing a fair and balanced summary. 
        Ensure the response maintains balance, captures key points, and distinguishes any opposing opinions. 
        Present the answer *short and concise*, phrased in a direct format without using phrases like "participants in the debate" or "in the debate."\\
        \\
        \#\#\# Input format\\
        - Debate topic: \{debate\_topic\}\\
        - Participant's responses:\\
          - Response 1: "\{response content\}" (Perspective: \{perspective 1\})\\
          - Response 2: "\{response content\}" (Perspective: \{perspective 2\})\\
          - ...\\
          - Response N: "\{response content\}" (Perspective: \{perspective N\})\\
        \\
        \#\#\# Output format\\
        A short and concise summary from the mediator’s perspective based on the discussion, phrased as a direct answer without reference to the debate structure or participants\\
        \\
        \#\#\# Inputs\\
        Debate topic: \{\texttt{debate\_topic}\}\\
        Participant's responses: \{\texttt{responses}\}\\
        \#\#\# Output\\
        Summary: 
        \vspace{-1mm}
    \end{tcolorbox}
    \vspace{-3mm}

    \caption{Prompt templates for generating sub-queries (top) and debate mediator (bottom) in Debate type questions. We reference the debate mediator prompt from \citet{liang-etal-2024-encouraging} to ensure that responses objectively aggregate and present diverse perspectives.}
    \label{fig:DEBATE_Prompt}
\end{figure*}
\clearpage

\twocolumn
\clearpage

\section{Implementation Details}

\label{sec:appendix_implementation_details}

\begin{figure*}[ht!]
    \centering
    \includegraphics[width=\textwidth]{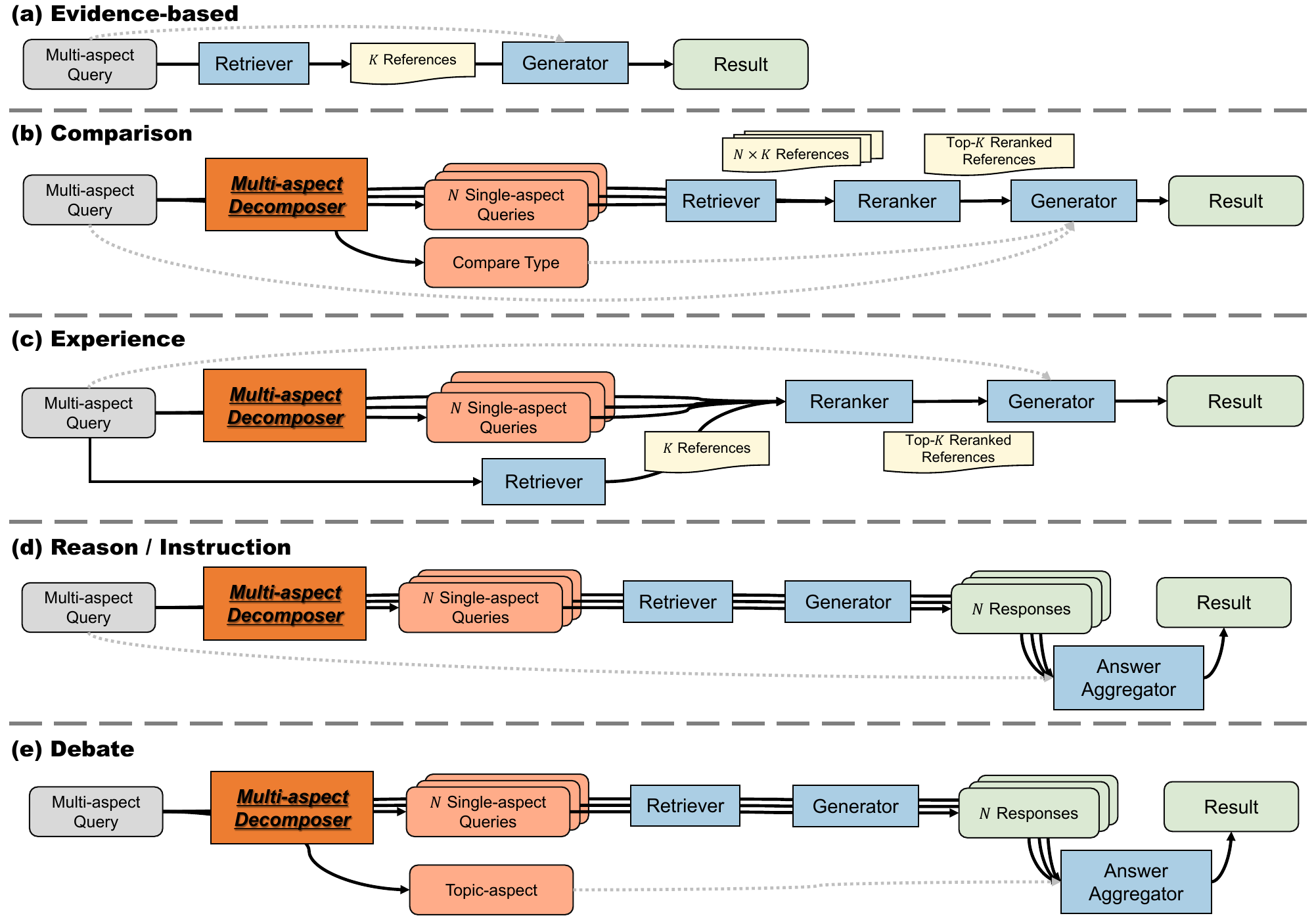}
    \caption {A detailed overview of \method.}
    \label{fig:detailed_overview}
\end{figure*}

All experiments were conducted using the NVIDIA A100 (80 GB) GPUs and the OpenAI API.

\subsection{LLM}

For all experiments involving open-source LLMs, we employ vLLM \cite{Kwon2023vLLM} to enable fast and memory-efficient inference.

\subsection{RAG}

To perform NFQA with our RAG-based QA system, the retriever selects five passages that are then provided to the generator as references. For Wikipedia-based tasks, we use BM25 on the Wikipedia corpus preprocessed by \citet{karpukhin-etal-2020-dense} as the external retrieval index.

\subsection{ \method}
A detailed overview of \method\ is shown in Figure \ref{fig:detailed_overview}.  

The reranker employed in \method\ is BGE-Reranker-Large\footnote{\url{https://huggingface.co/BAAI/bge-reranker-large}} \cite{Xiao2024bge_embedding}.

\subsection{LINKAGE Evaluation}
For LINKAGE evaluation, we adhere to the original settings: nucleus sampling (\texttt{top\_p} = 0.95), a maximum output length of 512 tokens, and a default temperature of 0.8. The temperature is reduced to 0.1 when annotating reference answers.

\subsection{Reference Answer Construction on Wiki-NFQA}
To build reference answers for the Wiki-NFQA dataset, we use three LLMs to capture diverse styles: (i) GPT-3.5-turbo-16k, (ii) Mistral-7B-Instruct-v0.2\footnote{\url{https://huggingface.co/mistralai/Mistral-7B-Instruct-v0.2}} \cite{Jiang2023Mistral-7B}, and (iii) Llama-3.1-8B-Instruct\footnote{\url{https://huggingface.co/meta-llama/Llama-3.1-8B-Instruct}}. 
Each model generates three high-quality responses, totaling nine reference answers. Additionally, we use GPT-4o-2024-08-06 \cite{OpenAI-2024-GPT-4o} to produce a single, superior reference answer that is distinct from the other nine.

\section{Detailed Analysis per Dataset}

We conduct a dataset-specific analysis to highlight \method ’s strengths across different challenge contexts.

\paragraph{NQ-NF, SQuAD-NF, and TriviaQA-NF}
These benchmarks feature open-ended non-factoid questions that often demand explanatory or elaborate answers. \method\ delivers substantial gains in both MRR and MPR, demonstrating its ability to generate detailed, on-topic responses. For instance, on SQuAD-NF using the Mistral-7B base model with a GPT-4o mini scorer, \method\ achieves an MRR of 0.7444—significantly outperforming the LLM-only (0.4222) and RAG-based (0.3817) baselines.

\paragraph{HotpotQA-NF and MuSiQue-NF}
These datasets require multi-hop reasoning, where answers must synthesize information from multiple passages. By decomposing questions into type-specific aspects, \method\ more effectively navigates these complex reasoning chains, yielding notable improvements in MPR compared to both LLM and RAG methods.

\paragraph{2WikiMultiHopQA-NF}
Although overall scores are lower on this particularly challenging dataset, \method\ still surpasses LLM-only and RAG baselines. This result underscores \method ’s robustness even in scenarios demanding extensive, multi-step inference.

\clearpage
\onecolumn

\section{Case Study}
\label{sec:appendix_case_study}
\centering
\renewcommand{\arraystretch}{1.3} 
\begin{longtable}{p{2.5cm} p{4cm} p{8cm}}
\toprule
\textbf{NFQ Type} & \textbf{Example of Question} & \textbf{Response of \method} \\ 
\midrule
\endfirsthead
\toprule
\textbf{NFQ Type} & \textbf{Example of Question} & \textbf{Response of \method} \\ 
\midrule
\endhead
\midrule
\multicolumn{3}{r}{\textit{Continued on next page}} \\
\midrule
\endfoot
\endlastfoot

Evidence-based & 
\raggedright  How does sterilisation help to keep the money flow even? & 
The term "sterilisation" is often used in economics to refer to the act of preventing currency or securities from affecting the monetary base due to interest rate differentials. When a central bank engages in open market operations to buy or sell securities, it may want to prevent the money flow from being disrupted by interest rate differentials in different markets.\newline \newline Sterilisation helps to keep the money flow even by offsetting the effect of open market operations on the monetary base. When a central bank buys securities, it injects cash into the economy and increases the monetary base. However, if this leads to an appreciation of the domestic currency, it may reduce the competitiveness of exports and increase the cost of imports. To prevent this, the central bank may sell an equivalent amount of foreign currency in the foreign exchange market, which reduces the monetary base and helps to offset the initial effect of the open market operation.\newline \newline Similarly, if a central bank sells securities, it decreases the monetary base and may lead to a depreciation of the domestic currency, which can make exports more competitive and imports more expensive. To prevent this, the central bank may buy an equivalent amount of foreign currency in the foreign exchange market, which increases the monetary base and helps to offset the initial effect of the open market operation.\newline \newline In summary, sterilisation is a tool used by central banks to keep the money flow even and maintain control over the monetary base, despite open market operations that may otherwise disrupt the flow of funds. It allows central banks to implement monetary policy while minimizing the impact on currency values and other macroeconomic variables. 
\\ \midrule

Comparison & 
\raggedright what is the difference between dysphagia and odynophagia & 

Dysphagia and odynophagia are two distinct conditions related to swallowing.\newline \newline Dysphagia is the medical term for difficulty swallowing. It can be caused by various conditions, including neurological disorders, stroke, cancer, or structural problems in the throat or esophagus. Symptoms of dysphagia include the sensation of food sticking in the throat or chest, coughing or choking during eating, and pain or discomfort while swallowing.\newline \newline On the other hand, odynophagia is the medical term for pain while swallowing. It is a symptom of an underlying condition, such as a sore throat, inflammation, or infection. The pain can be described as sharp or throbbing and may be worsened by swallowing solid or acidic foods. Odynophagia can also be a symptom of gastroesophageal reflux disease (GERD) or eosinophilic esophagitis.\newline \newline In summary, dysphagia is a condition that makes it difficult to swallow, while odynophagia is a symptom of pain while swallowing. A person with dysphagia may also experience pain while swallowing, but the primary issue is the difficulty in moving food from the mouth to the stomach. Conversely, a person with odynophagia primarily experiences pain while swallowing, but they may still be able to swallow relatively normally.
\\ \midrule

Experience & 
\raggedright What are some of the best Portuguese wines? & 

Based on the available references in the document, some Portuguese wines that are often considered excellent include:\newline 1. The reference 14098516 might be referring to Port Wine, a fortified wine produced in the Douro Valley region. Port Wine is known for its sweet and rich flavors and comes in various styles such as Tawny, Ruby, and White Port.\newline 2. Reference 5043240 could be referring to Vinho Verde, a light, slightly effervescent wine from the Vinho Verde region in the north of Portugal. It is typically low in alcohol and is often served chilled.\newline 3. The Douro region is also famous for its red wines, such as those referenced in 2866435. These wines are known for their robust flavors and are often blended from a variety of grape varieties.\newline 4. The reference 4891311 might be referring to Madeira Wine, another fortified wine from the Madeira Islands. Madeira Wine is known for its complex flavors and can range from sweet to dry.\newline 5. Lastly, the reference 309483 could be referring to D\~a Wines, which come from the D\~a region in central Portugal. These wines are known for their elegance and balance, with flavors of red fruit, herbs, and minerals.\newline \newline So, based on the given references, some of the best Portuguese wines include Port Wine, Vinho Verde, Douro Red Wines, Madeira Wine, and D\~a Wines.
\\ \midrule

Reason & 
\raggedright Kresy, which roughly was a part of the land beyond the so-called Curson Line, was drawn for what reason? & 

The Curzon Line refers to the boundary proposed by Lord Curzon, then Viceroy of India, in 1920, as a potential eastern border for a future independent Poland. Kresy, or Eastern Borderlands, was the area of land beyond this proposed border. It was drawn for various reasons, including historical, ethnic, and strategic considerations. The area had been part of the Polish state in the past, and there was a significant Polish population there. Ethnically, the region was diverse, with various groups including Poles, Ukrainians, Belarusians, and Jews. Strategically, the region was important due to its location near the borders of several neighboring countries and its rich natural resources.
\\ \midrule

Instruction & 
\raggedright How can you find a lodge to ask to be a member of? & 

To find a lodge to inquire about membership, consider the following steps:\newline \newline 1. Contact the Grand Lodge of your jurisdiction (state or country) for a list of lodges in your area. You can usually find this information on their website or by contacting them directly.\newline 2. Research the lodges on the list to learn more about their history, traditions, and meeting times. This can be done through their websites, social media pages, or by contacting them directly.\newline 3. Attend a few meetings as a visitor to get a feel for the lodge and its members. You may need to request an invitation from the lodge's secretary or master.\newline 4. Speak with current members about their experiences and the expectations for new members. This can help you determine if the lodge is a good fit for you.\newline 5. Consider your availability and the lodge's meeting schedule to ensure you can commit to attending meetings regularly.\newline \newline Remember, every lodge is unique, so it's essential to find one that aligns with your values and interests. Good luck in your search!\newline \newline (Note: The references provided do not contain information relevant to finding a lodge to ask to be a member of.)
\\ \midrule

Debate & 
\raggedright I Can See Your Voice, a reality show from South Korea, offers what kind of performers a chance to make their dreams of stardom a reality? & 

Based on the information provided in the references, "I Can See Your Voice" is a South Korean reality show that offers singers a chance to audition without revealing their appearance or singing voice to the judges initially. This format is designed to evaluate contestants based on their charisma, rhythm, and overall performance without any biases influenced by their physical appearance or singing voice. Therefore, the show provides an opportunity for talented singers who might not have been given a chance due to their looks or voice in their initial auditions to showcase their abilities and pursue their dreams of stardom.
\\ \bottomrule

\caption{Sample responses of \method\ across various non-factoid question types.}
\label{tab:case_study}
\end{longtable}

\end{document}